\documentclass[11pt]{article}

\usepackage[preprint]{acl}

\usepackage{times}
\usepackage{latexsym}
\usepackage{amsmath}
\usepackage{booktabs}
\usepackage{graphicx}
\usepackage{subcaption}
\usepackage{tikz}
\usepackage{multirow} 
\usepackage{microtype}
\usepackage{adjustbox}
\usepackage{xcolor}
\usepackage{colortbl}
\definecolor{posvstrong}{RGB}{255,120,120}    
\definecolor{posstrong}{RGB}{255,140,140}     
\definecolor{posmodstrong}{RGB}{255,160,160}  
\definecolor{posmod}{RGB}{255,180,180}        
\definecolor{posweak}{RGB}{255,215,215}       
\definecolor{posveryweak}{RGB}{235,235,235}   
\definecolor{posnearzero}{RGB}{245,245,245}   

\definecolor{negvstrong}{RGB}{120,120,255}    
\definecolor{negstrong}{RGB}{140,140,255}     
\definecolor{negmodstrong}{RGB}{160,160,255}  
\definecolor{negmod}{RGB}{180,180,255}        
\definecolor{negweak}{RGB}{215,215,255}       
\definecolor{negveryweak}{RGB}{235,235,235}   
\definecolor{negnearzero}{RGB}{245,245,245}   
\usepackage{hyperref}
\usepackage{stfloats}
\usepackage{placeins}
\usepackage{longtable}
\usepackage{amsmath}    
\usepackage{amssymb}    
\usepackage{algorithm}  
\usepackage{algpseudocode}  

\usepackage[textsize=tiny]{todonotes}
\usepackage{listings}
\title{Isotropy Cliffs: The Geometric Signature of Decision-Making in Large Language Models}

\author{
  Okan S. Coskun\textsuperscript{1}, Florian Rottach\textsuperscript{1}, Carsten Eickhoff\textsuperscript{1}, \bf William Rudman\textsuperscript{2}, \\
  \textsuperscript{1}University of Tübingen \quad
  \textsuperscript{2}The University of Texas at Austin \\[0.4em]
  \small Correspondence: \texttt{okan.coskun@guest.uni-tuebingen.de} \  
}

\begin{document}
\maketitle

\begin{abstract}

We investigate the geometry of decision-making in Multiple Choice Question Answering (MCQA) through the lens of isotropy. Analyzing five open-weight models across diverse datasets, we identify decision-critical transition layers characterized by a shift in isotropy, coinciding with a major representational change and the emergence of task-relevant clusters. We demonstrate that this synchronized geometric behavior is strongly correlated with downstream accuracy ($r\approx0.84$), demonstrating its relevance for successful decision-making. Furthermore, we show that this transition is robust to prompt variations and in-context learning (ICL), suggesting that it reflects a general mechanism of model behavior. Lastly, we verify the causal relevance of isotropy cliffs by intervening at the valley of the cliff (layer $L+1$) to prevent the sudden drop in isotropy. Increasing isotropy at this layer disrupts the formation of salient clusters in the embedding space and significantly reduces accuracy. Together, these findings reveal a consistent geometric transition associated with decision-making and provide insight into how task-relevant structure emerges in the representations of LLMs.
\end{abstract}

\section{Introduction}

\begin{figure}[!t]
\centering
\includegraphics[width=0.9\textwidth]{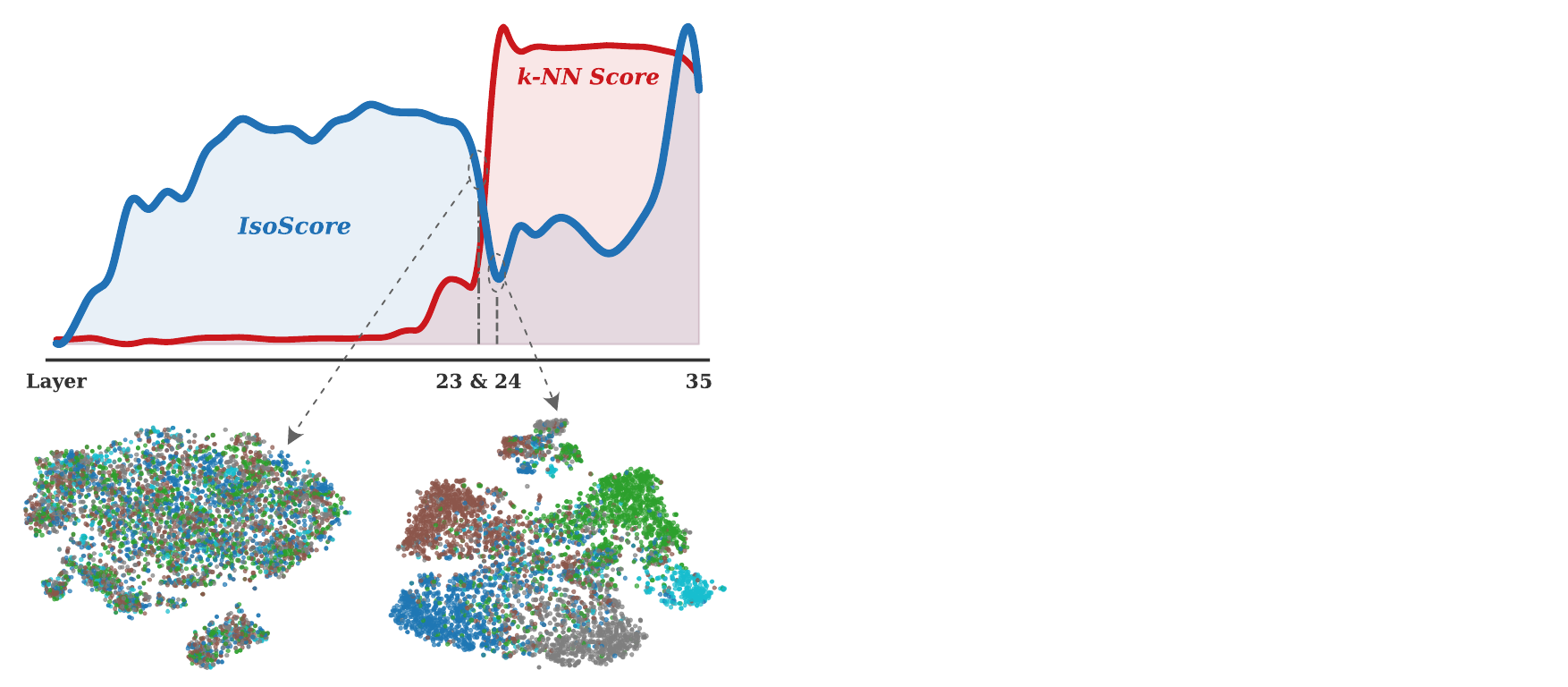}
\caption{\textbf{Dynamics of Representation Collapse and Cluster Formation.} Analysis of the Qwen3-8B model on the Head-QA dataset. The dual-axis plot tracks the IsoScore and $k$-NN Score, showing a critical transition between the layers 23--24. The t-SNE projection at layer 23 shows overlapping representations with high isotropy, while t-SNE projection at layer 24 shows the emergence of distinct class clusters following the structural collapse of the representation space.}
\label{fig:qwen3_iso_cluster}
\end{figure}

Several works have studied the geometry of Large Language Model (LLM) embeddings \cite{ghandeharioun2024patchscopesunifyingframeworkinspecting, zhou-etal-2024-alignment, ansuiniIntrinsicDimensionData2019}, how concepts are represented in LLM embeddings \cite{marks2024geometrytruthemergentlinear, turner_SteeringLanguage_2024} and how these representations evolve across layers \cite{liuSpectralInsightsDataOblivious2025}. Prior work finds that representations first undergo expansion, marked by an increase in intrinsic dimension in embedding space, followed by a gradual compression into a lower dimensional subspace \cite{ansuiniIntrinsicDimensionData2019}. This compression can be extreme, with even a one-dimensional space in later layers retaining enough information to solve a classification task \cite{rudman-etal-2023-outlier}. Together, these dynamics are thought to reflect distinct stages of reasoning and abstraction \cite{gardinazzi_PersistentTopological_2025}.
While these findings have improved our understanding of LLM representations, existing approaches primarily focus on local estimates of intrinsic dimensionality using TwoNN, which do not scale well to high-dimensional data~\citep{lid} and can have high variance~\citep{lid_variance}. In this paper, we study how isotropy (i.e. the uniformity of variance across dimensions) evolves across LLM layers. While \textit{isotropy} has been studied as a property of embedding spaces \cite{cai2021isotropy}, its role in the layer-wise evolution of representations and its connection to performance remains unclear. Our results show that isotropy values rise steadily through the network's early layers, peaking at critical layer $L$ before dropping sharply at layer $L+1$. We call this characteristic behavior the \textit{isotropy cliff}. Evaluated across five LLMs, four question answering tasks, four prompt variations, and in-context learning samples, we find that the \textit{steepness} of the isotropy cliff is a strong predictor of task accuracy, achieving Spearman correlations as high as 0.92 on Mistral-7B. The main contributions of this work are the following:

    \begin{itemize}
    \item We identify a consistent geometric transition in the middle-to-late layers in LLMs, characterized by a sharp drop in isotropy (\textit{isotropy cliff}), which coincides with the emergence of task-relevant cluster structure.

    \item We demonstrate that the steepness of this isotropy cliff is a strong predictor of downstream task accuracy, achieving Spearman correlations of up to $\rho = -0.92$, outperforming Intrinsic Dimensionality as a performance predictor. Further, we demonstrate that preventing the formation of isotropy cliffs through causal intervention significantly decreases model performance.

    \item We show that this geometric transition persists across prompt variations and in-context learning strategies, suggesting that it reflects an intrinsic mechanism of decoder-only model processing rather than an artifact of a specific prompt format or architecture.\textbf{}
\end{itemize}

\section{Related Work}

Recent work has moved beyond static analysis of model layers to characterize the dynamic evolution of internal representations. Several studies demonstrate the existence of distinct ``processing phases'' as information propagates through the network. Intrinsic Dimension (ID) has emerged as a primary tool for mapping these trajectories.  Recent work identifies a universal ``high-dimensional abstraction phase'' in intermediate layers, arguing that this expansion encodes complex linguistic dependencies before transferring to downstream tasks \cite{cheng_EmergenceHighDimensional_2025}. Others investigate how the geometry of token representations across layers relates to next-token prediction, using intrinsic dimension to show that high prediction loss is associated with representations that have a high intrinsic dimension \cite{viswanathan2025geometrytokensinternalrepresentations}. In the context of Multiple Choice Question Answering (MCQA), \citet{joshi2025geometrydecisionmakinglanguage} link these dimensional ID shifts to decision-making dynamics. The authors observe a ``hump-shaped'' ID profile where early expansion allows for exploration, while the subsequent compression in later layers coincides with the model becoming decisive about the final answer token. \citet{skean2025layer} analyze representation quality across layers and find that intermediate LLM layers often outperform the final layer on 32 MTEB tasks. Complementary to this perspective, we focus on the geometric organization of representations, analyzing how isotropy evolves across layers and relates to the emergence of task-aligned structure. 
\citet{liuSpectralInsightsDataOblivious2025} identify critical layers using CKA, where low CKA similarity between a layer and its successor indicate a significant change in representation. They further validate this by fine tuning and freezing individual layers, showing that the largest effects occur at layers with low CKA values. Similarly, \citet{lad2025remarkable} observe block-like CKA structure across model families and find that early and final layers are most sensitive to perturbations.

\begin{figure}[t]
    \centering
    \includegraphics[width=0.9\linewidth]{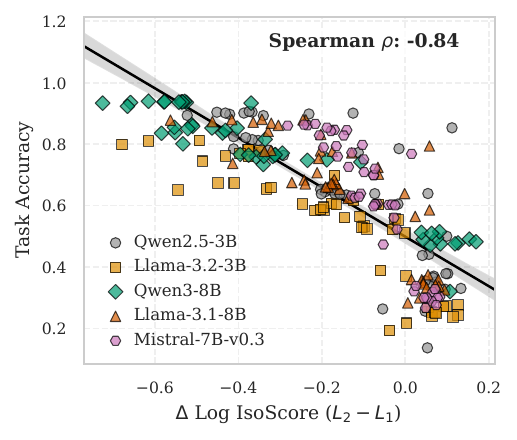}
    \caption{\textbf{Relationship between layer-wise isotropy dynamics and model performance.} Each data point represents a specific evaluation configuration (task, prompt format, and shot-count) for each of the five models at two consecutive respective layers.} 
        \label{fig:isotropy_accuracy}
\end{figure}

Parallel to the analysis of intrinsic dimensions of LLM embeddings, isotropy remains a critical yet debated property. Early pre-trained models were known to be anisotropic and have variance concentrated along only a small subset of dimensions \cite{ethayarajh_HowContextual_2019, razzhigaev2024shapelearninganisotropyintrinsic}. \citet{xiao_IsotropyContextualization_2023} demonstrate that contrastive learning objectives can effectively reduce this lack of isotropy (i.e., anisotropy) and cause variance to be more uniformly distributed. However, the relationship between isotropy and semantic structure is debated. While some studies show that isotropy can be beneficial \cite{bihani_LowAnisotropy_2021} in certain tasks such as retrieval \cite{kim-etal-2024-hil, karpukhin-etal-2020-dense, rottach2025topology}, recent studies have also indicated that isotropy is not compatible with clustering \cite{mickus-etal-2024-isotropy, ait-saada-nadif-2023-anisotropy}, demonstrating that isotropy as a desired property can be task dependent \cite{tsukagoshi2025redundancyisotropyintrinsicdimensionality}. Much of the debate on the role of isotropy stems from the variety of methods used to measure it, with different metrics sometimes yielding conflicting conclusions. Some metrics rely on the average cosine similarity of embeddings \cite{ethayarajh_HowContextual_2019} while others are based on spectral properties of embeddings \cite{mu2018allbutthetopsimpleeffectivepostprocessing}. \citet{rudmanIsoScoreMeasuringUniformity2022} argue that these traditional metrics are flawed, proposing IsoScore as a more  accurate way to measure isotropy.

\begin{figure*}[!t]
\centering

\includegraphics[width=\textwidth]{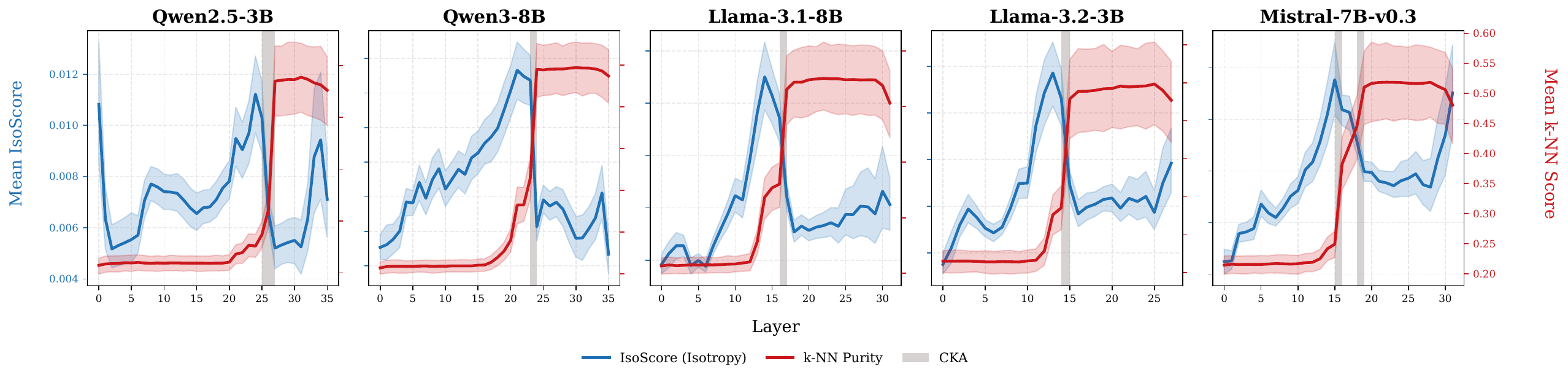}
\caption{\textbf{Layer-wise evolution of mean IsoScore and $k$-NN with 95\% Confidence Intervals (CI).} For each model, values are averaged across all evaluated datasets, prompt variants, and in-context learning strategies. Shaded regions indicate the layers of maximal representational change as identified by CKA (see Figure \ref{fig:cka_all}). We observe that clustering (red) and the drop in isotropy (blue) are precisely aligned within these zones.}
\label{fig:isotropy_all}
\end{figure*}

Prior work studying the geometry of LLM embeddings has not systematically accounted for the influence of prompting and ICL strategies on model geometry \cite{zhuo2024prosaassessingunderstandingprompt, loya-etal-2023-exploring, zhu2024promptrobustevaluatingrobustnesslarge}. Different prompting strategies can lead to substantial performance differences \cite{lyu_ProbabilitiesUnveiling_2024, pezeshkpour-hruschka-2024-large}. For instance, \citet{robinson_LeveragingLarge_2023} show that restructuring prompts from cloze-style formulations to explicit multiple-choice formats yields large and consistent performance gains across diverse datasets. 
More broadly, models are highly sensitive to seemingly superficial prompt variations \cite{mu2024navigatingpromptcomplexityzeroshot, raj2025semanticconsistencyassuringreliability}. Performance can vary dramatically across semantically equivalent prompt formats \cite{sclar2024quantifyinglanguagemodelssensitivity}, and even the presentation of answer options (e.g., bullet points vs.\ plain text) can significantly affect accuracy \cite{han2025effectselectionformatllm}. These formatting changes can also alter internal processing. \citet{wiegreffe2025answer} find that MCQA answer selection is driven by a sparse set of attention heads in a few middle layers, and that when options are labeled with unusual symbols, the answer is only resolved in later layers.
These findings suggest that prompt design acts as a highly influential factor in evaluation and their impact on embedding space geometry should be studied. By explicitly incorporating diverse prompt formulations and ICL settings, we account for this representational variance, which we show influences task accuracy and meaningfully change the geometric structure of model representations.


\begin{figure*}[!t]
\centering

\includegraphics[width=0.9\textwidth]{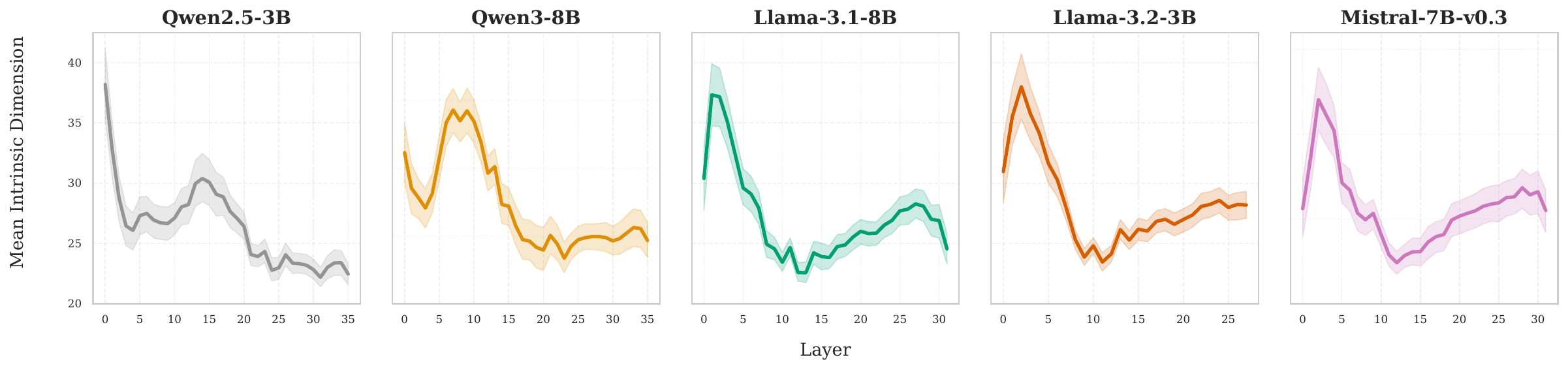}
\caption{\textbf{Layer-wise evolution of mean intrinsic dimension with 95\% confidence intervals (CI).} For each model, intrinsic dimension is averaged across all evaluated datasets, prompt variants, and in-context learning strategies, following the aggregation procedure in Figure \ref{fig:isotropy_all}.}
\label{fig:int_dim_all}
\end{figure*}

\section{Methodology}
\label{sec:methodology}

\vspace{-0.5em}

We examine decision making through the evolution of representation geometry across layers. Our analysis centers on changes in isotropy, using these transitions as an indicator of when task-relevant structure begins to emerge. To study this, we construct a layer-wise analysis framework that tracks: isotropy using IsoScore, intrinsic dimension using TwoNN, clustering using $k$-NN Purity scores and layer-wise similarity using CKA. Together, these measurements allow us to identify where and how representations reorganize into decision-critical states. To capture a wide range of task structures, we select four distinct multiple-choice benchmarks: ARC-Easy, OpenBookQA, Head-QA and MMLU-Pro , covering diverse domains, varying reasoning difficulties, and different numbers of answer choices. We evaluate Qwen2.5-3B, Qwen3-8B, Llama-3.1-8B, Llama-3.2-3B and Mistral-7B-v0.3, covering a range of different families, architectures and sizes. Detailed information on all datasets and models, including their respective sources, is provided in Appendix Table~\ref{tab:appendix_models} and \ref{tab:appendix_datasets}.

\paragraph{Model Inference.}  Following \citet{robinson_LeveragingLarge_2023}, we evaluate performance by examining the probability distribution over the final output logits. We select the token $\hat{y}$ with the highest probability from the valid answer set (e.g., $\{\text{``A''}, \text{``B''}, \dots\}$); if $\hat{y}$ matches the true label, the sample is recorded as correct. We also extract the internal representations of the model at this decision point. Specifically, let $\mathbf{H}^\ell$ denote the hidden state matrix (the residual stream) at layer $\ell$. We extract the final row of this matrix, $\mathbf{H}_T^\ell$, which corresponds to the $T$-th and final token of the input prompt. This vector serves as our sentence embedding:
\begin{equation}
    \mathbf{s}^\ell = \mathbf{H}_T^\ell \in \mathbf{R}^D
\end{equation}
where $D$ is the model's internal embedding dimension. By collecting these vectors across all $L$ layers for a dataset of $N$ samples, we construct an activation tensor of shape $(N, L, D)$. This tensor forms the empirical basis for our geometric analysis of the model's internal representations.

\subsection{Geometric and Similarity Measures}
\label{subsec:metrics}

We employ a suite of metrics to characterize the embedding space, distinguishing between global geometry, local separability, and representational similarity.
We characterize the point cloud using \textbf{isotropy}, which measures how uniformly the variance of the embeddings is distributed across all dimensions. After reorienting the data into the PCA basis, the isotropy defect is computed as the normalized Euclidean distance between the normalized variance vector $\hat{\Sigma}_D$ and the perfectly isotropic vector $\mathbf{1}$:

\begin{equation}
\delta(X) :=
\frac{\left\|\hat{\Sigma}_D-\mathbf{1}\right\|}
{\sqrt{2(n-\sqrt{n})}}.
\end{equation}

The isotropy defect is then linearly transformed into the final \textbf{IsoScore} \citep{rudmanIsoScoreMeasuringUniformity2022},
where a score of 1 indicates perfect uniformity, while 0 implies the variance is collapsed into a few dominant directions (anisotropy).
To assess task-relevant structure, we measure local cluster purity via the \textbf{$k$-NN Score} (with $k=5$). For a given embedding $\mathbf{e}_i$, let its $k$ nearest Euclidean neighbors be $\mathbf{n}_1, \ldots, \mathbf{n}_k$. The local purity is computed as:
\begin{equation}
    \text{k-NN}(\mathbf{e}_i) =
    \frac{1}{k} \sum_{j=1}^k
    \mathbf{I}(L(\mathbf{n}_j) = L(\mathbf{e}_i))
    \label{eq:local_purity}
\end{equation}
where $L(\cdot)$ yields the ground-truth label and
$\mathbf{I}$ is the indicator function. To obtain the overall k-NN Score (KS) for a specific layer $\ell$, we simply average this local score across all $N$ embeddings in that layer. A high KS signals that the model is effectively separating task-relevant classes in the representation space, a structural shift we later visualize using t-SNE \cite{JMLR:v9:vandermaaten08a}. We further compute the \textbf{Adjusted Rand Index (ARI)} between the unsupervised $k$-means clusters and two distinct label assignments, ground-truth labels (GT) and the model's own predictions (PD). Evaluating both allows us to distinguish structure associated with the dataset's ground-truth labels from structure that is also reflected in the model's own predictions. 

\textbf{Intrinsic Dimensionality (ID)} determines how many dimensions are needed to effectively represent a point cloud $X \in \mathbb{R}^n$. The manifold hypothesis states that high-dimensional data lies on a lower-dimensional manifold whose intrinsic dimension is smaller than the ambient dimension $n$ \cite{bengio2013representation}. \citet{Facco_2017} introduce the Two Nearest Neighbors estimator \textbf{TwoNN}, which approximates the intrinsic dimension by analyzing the distribution of the ratio $\mu$ between each point's second and first nearest neighbor distances. The intrinsic dimension is then estimated as

\begin{equation}
d=\frac{\log(1-F(\mu))}{\log(\mu)},
\end{equation}

where $F(\mu)$ denotes the cumulative distribution function of the nearest-neighbor distance ratios.

Finally, to track how information evolves across layers, we use \textbf{CKA} \citep{pmlr-v97-kornblith19a}, which is defined as

\begin{equation}
\mathrm{CKA}(K_1,K_2)=
\frac{\|K_1K_2\|_F^2}
{\|K_1\|_F\,\|K_2\|_F},
\end{equation} where $K_l=\tilde{\mathbf{X}}^{(l)T}\tilde{\mathbf{X}}^{(l)}$, $\tilde{\mathbf{X}}$ denotes the centered representation matrix, and $\|\cdot\|_F$ the Frobenius norm. Unlike the measures above, CKA allows us to compute the similarity between two consecutive layers by comparing their underlying geometric structures, invariant to rotation and scaling. 

\begin{center}

\includegraphics[
    width=0.9\columnwidth
]{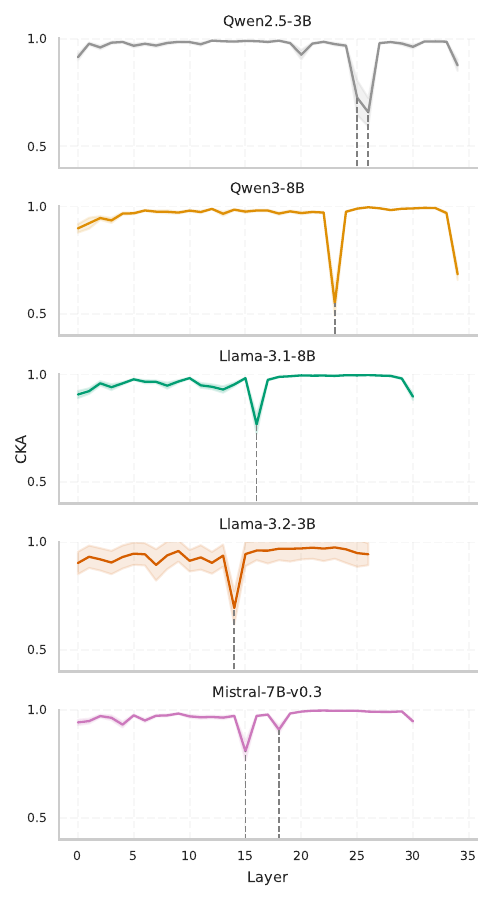}

\captionof{figure}[t]{
    \textbf{Layer-wise CKA trajectories}.
    Averaged across all evaluated datasets, prompt variants, and in-context learning strategies. Shaded regions
    indicate 95\% confidence intervals across tasks.
}
\label{fig:cka_all}

\end{center}

A low alignment score indicates a significant transformation between consecutive layers. Prior work has shown that such changes are not arbitrary, but rather concentrate at specific layers that play a critical role in model behavior \citep{liuSpectralInsightsDataOblivious2025}. In our setting, we show that these representational shifts coincide with changes in isotropy, allowing us to link these properties.

\vspace{-0.09em}

\section{Results}
Figure~\ref{fig:isotropy_all} shows how representational geometry evolves across layers in all five model families, plotting IsoScore (left $y$-axis, blue) and $k$-NN Score (right $y$-axis, red) for each layer.
The data reveals a consistent trend across all models. In the middle layers, IsoScore peaks, then suddenly drops. When isotropy drops, there is a sharp increase in $k$-NN Score, which indicates the emergence of clusters in activation space. This observation is consistent across all models and empirically validates the theoretical proofs that clustering and isotropy are negatively correlated~\citep{mickus-etal-2024-isotropy}. To further investigate this geometric transition, we analyze the representational similarity between consecutive layers using CKA. This allows us to track how the internal representation of information evolves or stabilizes as it passes through the model. 

CKA is computed between a layer $L$ and $L+1$ for all models, tasks, and averaged across prompt variations (Figure~\ref{fig:cka_all}). Since CKA tracks important changes in the models embeddings, we hypothesize that a drop in CKA will occur at the layer where IsoScore declines. Figure~\ref{fig:cka_all} shows that CKA scores are largely stable in the initial layers, followed by a notable decrease in similarity in middle layers. This representational change is observed \textbf{exactly at the isotropy cliff}. This relationship suggests that the decrease in isotropy characterizes a significant reorganization of the embedding space. Notably, this drop in CKA does not correspond with a significant change in intrinsic dimension.
The shaded, gray region in Figure~\ref{fig:isotropy_all} represents the layer where the drop from Figure~\ref{fig:cka_all} occurs. For the following analysis, we focus on the two transition layers for each model, chosen based on the jointly observed cliff in isotropy, $k$-NN Score, and CKA overlap, which together indicate a major reorganization of the representation geometry. Although the transition can span multiple neighboring layers, the comparison between two layers for each model reveals a substantial structural shift, as highlighted by the abrupt emergence of separable clusters (Figure~\ref{fig:qwen3_iso_cluster}) and furthermore, a correlation of isotropy with downstream task accuracy (Table~\ref{tab:global_transition_correlations_layers}).


\subsection{Emergence of Clusters}

\paragraph{Early decoding and clustering} We first quantify the emergence of class structure at the isotropy cliff by measuring clustering quality at the transition layers $L$ and $L+1$. Table~\ref{tab:silhouette_scores} reports Silhouette scores on the full-dimensional representations, together with early decoding accuracy and ARI against the ground-truth answer labels. Across all five models, the representations show little to no class separation at layer $L$, with Silhouette scores ranging from $-0.02$ to $0.04$, early decoding from $0.26$ to $0.58$, and ARI from $0.08$ to $0.31$. In contrast, at layer $L+1$, all three measures increase substantially.

This change is accompanied by a corresponding emergence of answer information in the representations, as measured by early decoding. While the representations at layer $L$ contain limited information that can be used to distinguish the answers, decoding performance increases sharply after the transition at $L+1$. Together, the clustering and decoding results indicate that the isotropy cliff marks a transition at which the representations become increasingly organized.

\begin{figure*}[!t]
\centering

\includegraphics[width=0.85\textwidth]{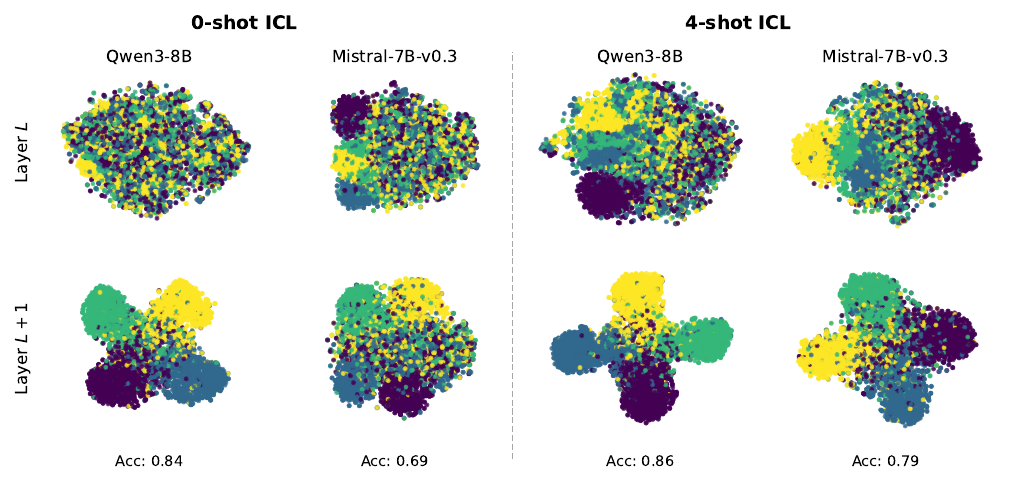}

\caption{
\textbf{Clustering analysis of model activations on OpenBookQA using prompt format 0.} Low-dimensional $t$-SNE projections of the residual stream show representations grouped by ground-truth answer choices. This alignment demonstrates behavioral clustering \cite{rimsky-etal-2024-steering}, where semantic distinctions between choices are reflected in the geometry of the latent space.}

\label{fig:tsne_chosen_main}
\end{figure*}

\paragraph{2D Projections Reveal Task-Relevant Clustering} To qualitatively visualize this representational reorganization, we project the same hidden representations into two dimensions using $t$-SNE. To ensure consistency across analyses, the projections are computed from the same hidden representations used in the previous experiments. For each input example, the corresponding representation is projected into two dimensions and labeled according to its ground-truth answer. If the model internally captures the distinction between the answer choices, representations associated with the same label are expected to lie closer in vector space than representations associated with different labels, which is also referred to as behavioral clustering \cite{rimsky-etal-2024-steering}. Figure~\ref{fig:tsne_chosen_main} shows Qwen3-8B and Mistral-7B-v0.3 at the isotropy cliff layers $L$ and $L+1$ under 0-shot and 4-shot in-context learning. At $L$, representations of different answer classes remain largely intermixed, whereas at $L+1$ they reorganize into more distinct clusters aligned with the ground-truth labels.

\begin{table}[t]
\centering
\caption{
Clustering quality and early decoding accuracy at isotropy cliff layers $L$ and $L+1$ on Arc-Easy using prompt format 2 and 4-shots. Silhouette is computed on the full-dimensional embeddings with the ground-truth answer letters as labels. ARI is computed from $k$-means ($k=4$) against the ground truth. Early Decoding Acc ($\text{Acc}_{\text{early}}$) indicates the accuracy obtained when decoding directly from the intermediate layer representations.
}
\label{tab:silhouette_scores}
\small
\renewcommand{\arraystretch}{1.25}
\begin{adjustbox}{width=\columnwidth}
\begin{tabular}{l rr rr rr}
\toprule
& \multicolumn{2}{c}{\textbf{Silhouette}}
& \multicolumn{2}{c}{\textbf{$\text{Acc}_{\text{early}}$}}
& \multicolumn{2}{c}{\textbf{ARI}} \\
\cmidrule(lr){2-3} \cmidrule(lr){4-5} \cmidrule(lr){6-7}
\textbf{Model}
& $L$ & $L{+}1$
& $L$ & $L{+}1$
& $L$ & $L{+}1$ \\
\midrule
Qwen2.5-3B       & $-0.02$ & 0.23 & 0.26 & 0.68 & 0.19 & 0.78 \\
Qwen3-8B         & 0.02    & 0.27 & 0.33 & 0.96 & 0.16 & 0.88 \\
Llama-3.1-8B     & 0.00    & 0.11 & 0.47 & 0.89 & 0.25 & 0.62 \\
Llama-3.2-3B     & $-0.01$ & 0.07 & 0.29 & 0.71 & 0.08 & 0.39 \\
Mistral-7B-v0.3  & 0.04    & 0.10 & 0.58 & 0.78 & 0.31 & 0.61 \\
\bottomrule
\end{tabular}
\end{adjustbox}
\end{table}

The final row reports the downstream task accuracy. Models exhibiting clearer and more compact post-transition cluster formation generally achieve higher task accuracy, whereas models with weaker or less distinct cluster separation tend to obtain lower performance. This pattern persists across both models and ICL settings and qualitatively reflects the sharp increase in clustering metrics reported in Table~\ref{tab:silhouette_scores}. 

\begin{table*}[t]
\centering
\caption{
Spearman correlations ($\rho$) between transition metrics and task accuracy. 
$\Delta$LI denotes the change in log isotropy from the pre-transition state ($L$) to the post-transition state ($L+1$). 
KS and CKA correspond to post-transition and pre-transition states, respectively.
}
\label{tab:global_transition_correlations_layers}
\small
\renewcommand{\arraystretch}{1.25}
\begin{adjustbox}{width=0.9\textwidth}
\begin{tabular*}{\textwidth}{l @{\extracolsep{\fill}} cccccc}
\toprule
\textbf{Model} 
& \textbf{Transition Layers ($L \rightarrow L+1$)}
& \textbf{$\Delta$LI--Acc} 
& \textbf{$\Delta$LI--KS$_2$} 
& \textbf{$\Delta$LI--CKA$_1$} 
& \textbf{$\Delta$LI--$\Delta$ID} 
& \textbf{$\Delta$ID--Acc} \\
\midrule
Qwen2.5-3B    & Layer 26 $\rightarrow$ Layer 27 & \cellcolor{negmodstrong}-0.78 & \cellcolor{negmodstrong}-0.75 & \cellcolor{posmodstrong}0.72  & \cellcolor{negweak}-0.46  & \cellcolor{posveryweak}0.31 \\
Llama-3.2-3B  & Layer 14 $\rightarrow$ Layer 15 & \cellcolor{negvstrong}-0.92   & \cellcolor{negvstrong}-0.91   & \cellcolor{posstrong}0.85     & \cellcolor{posnearzero}0.02   & \cellcolor{posnearzero}0.00 \\
Qwen3-8B      & Layer 23 $\rightarrow$ Layer 24 & \cellcolor{negvstrong}-0.90   & \cellcolor{negstrong}-0.89    & \cellcolor{posmodstrong}0.78  & \cellcolor{posnearzero}0.15   & \cellcolor{negnearzero}-0.09 \\
Llama-3.1-8B  & Layer 16 $\rightarrow$ Layer 17 & \cellcolor{negstrong}-0.82    & \cellcolor{negstrong}-0.86    & \cellcolor{posstrong}0.86     & \cellcolor{posnearzero}0.17   & \cellcolor{negveryweak}-0.25 \\
Mistral-7B-v0.3    & Layer 18 $\rightarrow$ Layer 19 & \cellcolor{negvstrong}-0.92   & \cellcolor{negvstrong}-0.93    & \cellcolor{posvstrong}0.93    & \cellcolor{negweak}-0.49  & \cellcolor{posveryweak}0.40 \\
\midrule
Global        & Aggregate Across Models        & \cellcolor{negstrong}-0.84    & \cellcolor{negstrong}-0.83    & \cellcolor{posstrong}0.81     & \cellcolor{negnearzero}-0.10  & \cellcolor{posnearzero}0.03 \\
\bottomrule
\end{tabular*}
\end{adjustbox}
\end{table*}


\subsection{Predictive Power \& Causal Validation}
\label{sec:re}

\paragraph{Isotropy cliffs track model performance} Table~\ref{tab:global_transition_correlations_layers} reports the Spearman correlations between the change in log-IsoScore ($\Delta$LI), Intrinsic Dimensionality ($\Delta$ID), $k$-NN Score (KS), CKA similarity at the pre-transition layer, and task accuracy for the decision-critical layer pairs identified across five model architectures. We find that our isotropy cliff metric is the best predictor of model performance. A strong negative correlation between $\Delta$LI and both KS and accuracy demonstrates that a drop in isotropy is fundamentally associated with an increase in both clustering quality and task performance. The relationship between $\Delta$LI and CKA similarity is positive for all models ($\rho=0.72$–$0.93$), meaning that larger drops in isotropy in the middle layers coincide with significant representational shifts between the adjacent layers. Since lower CKA values indicate greater representational change, this positive correlation reflects the fact that the geometric phase transition is accompanied by a massive restructuring of the embedding manifold. These results provide strong evidence that the concentration in anisotropy is precisely the point where the model undergoes a structural reorganization that enables improved performance and accuracy.

\paragraph{Preventing the isotropy cliff hurts model performance}
In order to test whether isotropy cliffs are causally related to model performance, we collect hidden representations from 4,000 OpenBookQA samples and apply the all-but-the-top algorithms proposed by~\citet{mu2018allbutthetopsimpleeffectivepostprocessing} to increase isotropy by removing a small set of principal components in the embedding space that cause representations to be anisotropic. We then evaluate the effect of this intervention on model performance using 1,000 additional unseen samples from OpenBookQA with prompt format 0. Removing the top-10 principal components corresponds to approximately 0.05--1\% of the embedding dimension depending on the model. Several works have shown that increasing isotropy with all-but-the-top leads to improved model performance~\cite{mu2018allbutthetopsimpleeffectivepostprocessing}. Figure~\ref{fig:isotropy_accuracy} shows the impact of increasing isotropy for layers in $[L-2, L-1, L, L+1]$. Increasing (or decreasing isotropy by subtracting) does not significantly change performance for models at layers $L-2, L-1$, or $L$, with a drop as low as 1\% for some models and even small performance increases for Llama-3.2. When we intervene at layer $L+1$, preventing the isotropy cliff from occurring, we see a significant drop in performance (up to $~40\%$). Importantly, this performance degradation is accompanied by a pronounced disruption of the task-relevant cluster structure. As shown in Table~\ref{tab:intervened_window_stacked}, accuracy at $L+1$ remains very low, with ARI collapsing to values close to zero across models. This shows that preventing the isotropy cliff disrupts the reorganization of representations initially observed at the transition. Notably, this disruption occurs only at $L+1$, the layer of the isotropy cliff.

Interestingly, Qwen3-8B displays notable resilience, experiencing only a minor decrease in final task accuracy despite the intervention. As detailed in Appendix~\ref{tab:intervened_window_stacked}, intervening at $L+1$ initially eliminates cluster formation and sharp transition in Qwen3-8B just as in other models. However it manages to progressively reorganize its representation manifold and re-establishes it in subsequent layers (Appendix \ref{fig:qwen3_clean_recovery}). While the immediate post-cliff transition remains the single most impactful decision site across architectures, this recovery capability highlights model-specific variations in representation resilience.

\section{Discussion and Conclusion}

\textbf{ID vs. Isotropy.} Our findings provide a new geometric lens on the ``processing phases'' previously identified through Intrinsic Dimension (ID). While ID research often describes a hump-shaped profile expansion followed by compression \cite{joshi2025geometrydecisionmakinglanguage}, investigating this through isotropy provides an clearer picture of how the global organization of the space changes. By measuring isotropy, we find a gradual increase in IsoScore that mirrors the gradual increase in intrinsic dimension. However, where intrinsic dimension gradually decrease to form a ``hunchback'' pattern, we observe a sudden, steep decline in isotropy, indicating an abrupt reorganization of the embedding space. Not only does this sudden decrease in isotropy align better with the sudden emergence of task-relevant clusters, but the intensity of these isotropy cliffs proves to be a superior predictor of downstream accuracy than intrinsic dimension.

\begin{figure}[t]
    \centering
    \includegraphics[width=0.9\linewidth]{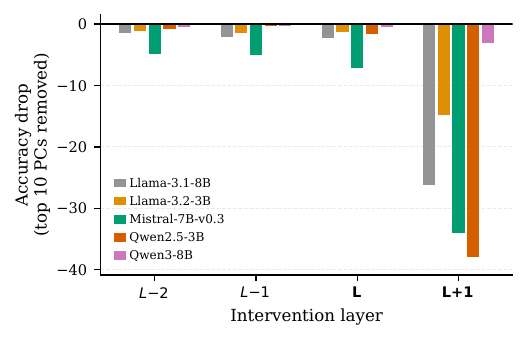}
   \caption{\textbf{Effect of top-10 principal-component removal around the isotropy cliff.}
 $L \rightarrow L+1$ denotes the isotropy cliff transition. Accuracy is reported relative to the corresponding baseline. Intervening at $L+1$, the layer associated with the isotropy drop, results in the largest decrease in accuracy across all models.}
        \label{fig:isotropy_accuracy}
\end{figure}

\textbf{From Anisotropy to Concentration.} Our findings align with previous work that finds LLM embeddings to be highly anisotropic \cite{timkey2021barkbiteroguedimensions,cai2021isotropy}. Early layers tend to be the least isotropic with IsoScore values of $(\sim 0.004)$. While isotropy increases in middle layers, peaks in IsoScore values remain extremely low $(\sim 0.012)$. In the layers before the isotropy peak, $k$-NN scores are near 0 indicating that no task-specific structure is present in the embedding space as points are not separable by their class label. However, we find a universal signature in LLM geometry that persists across different tasks, prompting strategies and model families: isotropy cliffs. These isotropy cliffs mark a sudden drop in isotropy and are accompanied by an emergence of task-specific clustering. We find that the steepness of this cliff is a strong predictor of model performance. In particular, we show that our isotropy cliff metric is a more reliable predictor than either intrinsic dimension or CKA. Interestingly, our isotropy cliff corresponds exactly with a sudden decrease in CKA followed by a return to near perfect values of CKA indicating that sudden drop in isotropy characterizes a two-stage processing in the geometry of LLM embeddings. 

In this work, we use isotropy to study how representations evolve across intermediate LLM layers. Models exhibit a distinct isotropy cliff that emerges at the same location regardless of the task, ICL strategy, or prompt. At the isotropy cliff, representations undergo a sharp drop in isotropy and reorganize into task-relevant clusters. The steepness of this drop predicts downstream accuracy better than alternative geometric measures such as intrinsic dimension or CKA. We causally validate the importance of the isotropy cliff by preventing it from forming at layer $L+1$ and show that this disrupts cluster formation and reduces accuracy. We believe that isotropy cliffs can serve as a practical tool for locating critical layers in LLMs and can help build a better understanding of how intermediate layer-wise geometry shapes model performance.

\section{Limitations}

\textbf{Identification of Transition Layers.}
The identification of the isotropy cliff currently relies on inspecting the layer-wise geometric profiles and subsequently defining the transition based on the largest change in isotropy together with the corresponding change in CKA. While this procedure provides a consistent basis for identifying the transition layers studied in this work, it is not yet formulated as a fully deterministic criterion that can be applied independently of visual inspection. A more systematic procedure for automatically identifying transition layers would strengthen the reproducibility of the analysis and provide a principled framework for comparing geometric transitions across models with different layer depths and representation dynamics.

\textbf{Locality of the Layer-wise Analysis.}
Our analysis focused only on pairs of adjacent layers, representing the minimum requirement needed to characterize a geometric transition. This localized view was already sufficient to reveal substantial representational restructuring and the emergence of task-relevant cluster formation. At the same time, more complex information may evolve across broader spans of layers. Models such as Qwen2.5-3B and Mistral-7B-v0.3, for example, undergo multiple representational shifts that are visible in both the CKA similarity drops and the fluctuations within their isotropy profiles. This suggests that for some models it may involve several stages of restructuring rather than a single isolated transition. We could therefore investigate broader layer spans and finer-grained layer dynamics in order to better understand how these geometric structures evolve throughout the network.

\textbf{Model and Task Scope.}
Beyond layer depth, expanding the scope of this analysis will be essential for testing the universality of these patterns. While we demonstrated consistent results across five model families and several diverse benchmarks, moving toward a wider variety of architectures and more complex tasks will provide a more comprehensive picture. Specifically, transitioning from the structured evaluation of multiple-choice question answering to the more fluid landscape of open-ended generation will be a critical next step.

\textbf{Qwen3 recovery} The recovery behavior observed in Qwen3-8B raises the question of why some architectures are able to compensate for the disruption of the isotropy cliff while others are not. A detailed comparison of the layer-wise representational dynamics across model architectures could help identify the mechanisms underlying this difference. In particular, it would be interesting to investigate whether differences in architectural design, depth, attention patterns, or the organization of information across layers allow Qwen3-8B to reconstruct task-relevant structure after the intervention. Understanding this recovery mechanism could provide further insight into how different decoder-only architectures distribute decision-relevant computation across their layers.

\bibliography{literatur/ref}

\appendix
    \section{Evaluation Prompt Formats}
\label{appendix:prompt_formats}

To evaluate the sensitivity of internal representations and model performance to prompt structure, we introduced four systematically varied prompt formats. The formats preserve the same semantic content and answer space while altering surface-level syntax, spacing, delimiters, capitalization, and structural organization.

The goal was to investigate whether even minor formatting perturbations, which are semantically irrelevant to humans, produce measurable differences in model behavior, representation geometry, or downstream accuracy.

The prompt formats range from minimal formatting deviations (Formats 0 and 3) to substantially different structural organizations (Formats 1 and 2). This allows us to analyze both subtle and strong prompt-induced distribution shifts.

The exact templates used, instantiated with an example question, are detailed below.

\subsection{Prompt Format 0: Standard Block Layout}
This format represents a classic, human-readable structured layout utilizing explicit category prefixes (Question, Choices, Answer) separated by clean vertical spacing and newlines.
\begin{lstlisting}
Question: What do plants need to make food?

Choices:
A: water and sunlight
B: only water
C: only sunlight
D: only air

Answer:
\end{lstlisting}

\subsection{Prompt Format 1: Continuous Prose / Inline Layout}
This format strips away formal structural blocks entirely, embedding the question, choices, and answer prompt into a single flowing narrative paragraph with parenthesis style separators. This evaluates the model's capacity to extract information when presented as standard conversational prose.
\begin{lstlisting}
What do plants need to make food? Choose from: (A) water and sunlight, (B) only water, 
(C) only sunlight, (D) only air. The answer is
\end{lstlisting}

\subsection{Prompt Format 2: Highly Delimited / Machine-Readable Layout}
This format capitalizes key identifiers and introduces synthetic semantic separators (double pipes \texttt{||} and semicolons \texttt{;}), mimicking structured configurations often used in programmatic data parsing. This tests the model's adaptability to strict, non-natural syntax rules.
\begin{lstlisting}
QUESTION: What do plants need to make food? || CHOICES: A=water and sunlight; 
B=only water; C=only sunlight; D=only air || ANSWER:
\end{lstlisting}

\subsection{Prompt Format 3: Compact Block Layout}
This format preserves similar phrasing and explicit prefixes compared to format 0, but eliminates vertical line breaks, compressing the choices onto a tighter vertical block to test the model's sensitivity to whitespace layout.
\begin{lstlisting}
Question: What do plants need to make food?
Choices: A: water and sunlight
B: only water
C: only sunlight
D: only air
Answer:
\end{lstlisting}
    \clearpage
\section{MCQA Results}
\begin{center}
\captionof{table}{Detailed Model Accuracy (\%) and IsoScore Delta in parentheses. For each format, the top row shows accuracy and the bottom row shows the IsoScore Delta.}
\label{tab:emnlp_fullwidth_stacked}
\scriptsize
\setlength{\tabcolsep}{5.0pt}
\begin{tabular}{ll ccc ccc ccc ccc}
\toprule
\textbf{Model} & \textbf{Format} & \multicolumn{3}{c}{\textbf{MMLU-Pro}} & \multicolumn{3}{c}{\textbf{OpenBookQA}} & \multicolumn{3}{c}{\textbf{ARC-Easy}} & \multicolumn{3}{c}{\textbf{Head-QA}} \\
\cmidrule(lr){3-5} \cmidrule(lr){6-8} \cmidrule(lr){9-11} \cmidrule(lr){12-14}
& & 0-s & 2-s & 4-s & 0-s & 2-s & 4-s & 0-s & 2-s & 4-s & 0-s & 2-s & 4-s \\
\midrule
\multirow{8}{*}{Qwen2.5-3B} 
 & \multirow{2}{*}{0} & 33.1 & 26.4 & 37.8 & 77.3 & 81.2 & 81.5 & 89.8 & 90.1 & 90.4 & 63.5 & 64.5 & 65.3 \\
 & & {\scriptsize (+.13)} & {\scriptsize (-.02)} & {\scriptsize (+.01)} & {\scriptsize (-.26)} & {\scriptsize (-.50)} & {\scriptsize (-.47)} & {\scriptsize (-.39)} & {\scriptsize (-.47)} & {\scriptsize (-.35)} & {\scriptsize (-.06)} & {\scriptsize (-.03)} & {\scriptsize (-.03)} \\
 \cmidrule(lr){2-14}
 & \multirow{2}{*}{1} & 25.7 & 35.5 & 33.8 & 60.5 & 77.7 & 78.5 & 85.3 & 90.1 & 89.4 & 50.0 & 63.7 & 63.8 \\
 & & {\scriptsize (+.05)} & {\scriptsize (+.15)} & {\scriptsize (+.06)} & {\scriptsize (+.05)} & {\scriptsize (-.43)} & {\scriptsize (-.39)} & {\scriptsize (+.09)} & {\scriptsize (-.38)} & {\scriptsize (-.32)} & {\scriptsize (+.01)} & {\scriptsize (-.04)} & {\scriptsize (-.05)} \\
 \cmidrule(lr){2-14}
 & \multirow{2}{*}{2} & 13.7 & 36.5 & 35.1 & 43.8 & 76.4 & 78.0 & 63.9 & 89.3 & 89.7 & 44.1 & 63.8 & 64.6 \\
 & & {\scriptsize (+.02)} & {\scriptsize (+.15)} & {\scriptsize (+.02)} & {\scriptsize (+.02)} & {\scriptsize (-.42)} & {\scriptsize (-.45)} & {\scriptsize (+.03)} & {\scriptsize (-.47)} & {\scriptsize (-.43)} & {\scriptsize (+.04)} & {\scriptsize (-.05)} & {\scriptsize (-.04)} \\
 \cmidrule(lr){2-14}
 & \multirow{2}{*}{3} & 32.8 & 37.9 & 37.0 & 78.5 & 82.1 & 82.5 & 90.1 & 90.7 & 90.5 & 63.9 & 65.1 & 65.7 \\
 & & {\scriptsize (+.09)} & {\scriptsize (+.06)} & {\scriptsize (-.01)} & {\scriptsize (-.08)} & {\scriptsize (-.49)} & {\scriptsize (-.44)} & {\scriptsize (-.29)} & {\scriptsize (-.46)} & {\scriptsize (-.44)} & {\scriptsize (-.04)} & {\scriptsize (-.03)} & {\scriptsize (-.03)} \\
\midrule
\multirow{8}{*}{Qwen3-8B} 
 & \multirow{2}{*}{0} & 47.9 & 32.0 & 49.1 & 83.6 & 85.5 & 86.3 & 93.6 & 93.8 & 93.6 & 74.9 & 75.9 & 76.6 \\
 & & {\scriptsize (+.13)} & {\scriptsize (+.11)} & {\scriptsize (+.15)} & {\scriptsize (-.58)} & {\scriptsize (-.55)} & {\scriptsize (-.51)} & {\scriptsize (-.66)} & {\scriptsize (-.58)} & {\scriptsize (-.52)} & {\scriptsize (-.19)} & {\scriptsize (-.36)} & {\scriptsize (-.31)} \\
 \cmidrule(lr){2-14}
 & \multirow{2}{*}{1} & 47.5 & 51.0 & 48.2 & 81.6 & 84.5 & 85.2 & 93.4 & 93.7 & 93.7 & 74.2 & 76.5 & 76.1 \\
 & & {\scriptsize (+.08)} & {\scriptsize (+.06)} & {\scriptsize (+.17)} & {\scriptsize (-.34)} & {\scriptsize (-.43)} & {\scriptsize (-.40)} & {\scriptsize (-.37)} & {\scriptsize (-.54)} & {\scriptsize (-.53)} & {\scriptsize (-.11)} & {\scriptsize (-.31)} & {\scriptsize (-.34)} \\
 \cmidrule(lr){2-14}
 & \multirow{2}{*}{2} & 47.2 & 51.3 & 48.9 & 80.1 & 83.6 & 85.3 & 92.4 & 93.8 & 93.8 & 73.4 & 76.4 & 76.7 \\
 & & {\scriptsize (+.06)} & {\scriptsize (+.04)} & {\scriptsize (+.04)} & {\scriptsize (-.53)} & {\scriptsize (-.44)} & {\scriptsize (-.46)} & {\scriptsize (-.67)} & {\scriptsize (-.54)} & {\scriptsize (-.58)} & {\scriptsize (-.34)} & {\scriptsize (-.38)} & {\scriptsize (-.40)} \\
 \cmidrule(lr){2-14}
 & \multirow{2}{*}{3} & 47.6 & 51.5 & 49.5 & 83.8 & 85.0 & 85.8 & 93.4 & 94.0 & 93.9 & 74.8 & 76.0 & 77.1 \\
 & & {\scriptsize (+.12)} & {\scriptsize (+.08)} & {\scriptsize (+.08)} & {\scriptsize (-.55)} & {\scriptsize (-.52)} & {\scriptsize (-.51)} & {\scriptsize (-.73)} & {\scriptsize (-.62)} & {\scriptsize (-.53)} & {\scriptsize (-.23)} & {\scriptsize (-.32)} & {\scriptsize (-.29)} \\
\midrule
\multirow{8}{*}{Llama-3.1-8B} 
 & \multirow{2}{*}{0} & 35.9 & 28.4 & 35.9 & 70.7 & 77.2 & 77.7 & 84.2 & 87.9 & 88.0 & 65.4 & 67.2 & 68.5 \\
 & & {\scriptsize (+.08)} & {\scriptsize (+.01)} & {\scriptsize (+.04)} & {\scriptsize (-.21)} & {\scriptsize (-.13)} & {\scriptsize (-.17)} & {\scriptsize (-.33)} & {\scriptsize (-.24)} & {\scriptsize (-.25)} & {\scriptsize (-.16)} & {\scriptsize (-.17)} & {\scriptsize (-.17)} \\
 \cmidrule(lr){2-14}
 & \multirow{2}{*}{1} & 30.0 & 34.2 & 33.3 & 58.6 & 75.4 & 77.8 & 79.4 & 87.2 & 88.1 & 56.5 & 66.1 & 67.2 \\
 & & {\scriptsize (+.03)} & {\scriptsize (+.07)} & {\scriptsize (+.05)} & {\scriptsize (+.06)} & {\scriptsize (-.21)} & {\scriptsize (-.20)} & {\scriptsize (+.06)} & {\scriptsize (-.36)} & {\scriptsize (-.34)} & {\scriptsize (+.02)} & {\scriptsize (-.19)} & {\scriptsize (-.18)} \\
 \cmidrule(lr){2-14}
 & \multirow{2}{*}{2} & 32.6 & 33.8 & 34.5 & 63.9 & 70.3 & 72.5 & 83.2 & 86.2 & 86.8 & 62.4 & 66.0 & 66.8 \\
 & & {\scriptsize (+.09)} & {\scriptsize (+.08)} & {\scriptsize (+.04)} & {\scriptsize (-.00)} & {\scriptsize (-.06)} & {\scriptsize (-.07)} & {\scriptsize (-.20)} & {\scriptsize (-.30)} & {\scriptsize (-.32)} & {\scriptsize (-.14)} & {\scriptsize (-.18)} & {\scriptsize (-.17)} \\
 \cmidrule(lr){2-14}
 & \multirow{2}{*}{3} & 37.5 & 35.9 & 35.1 & 73.8 & 77.7 & 78.0 & 87.8 & 87.3 & 88.0 & 67.5 & 67.1 & 68.4 \\
 & & {\scriptsize (+.05)} & {\scriptsize (+.01)} & {\scriptsize (+.04)} & {\scriptsize (-.41)} & {\scriptsize (-.34)} & {\scriptsize (-.32)} & {\scriptsize (-.56)} & {\scriptsize (-.46)} & {\scriptsize (-.42)} & {\scriptsize (-.27)} & {\scriptsize (-.24)} & {\scriptsize (-.24)} \\
\midrule
\multirow{8}{*}{Llama-3.2-3B} 
 & \multirow{2}{*}{0} & 25.4 & 21.7 & 26.9 & 60.5 & 65.4 & 65.9 & 74.5 & 77.7 & 78.1 & 53.4 & 56.1 & 56.4 \\
 & & {\scriptsize (+.06)} & {\scriptsize (+.00)} & {\scriptsize (+.05)} & {\scriptsize (-.25)} & {\scriptsize (-.33)} & {\scriptsize (-.32)} & {\scriptsize (-.49)} & {\scriptsize (-.35)} & {\scriptsize (-.38)} & {\scriptsize (-.10)} & {\scriptsize (-.15)} & {\scriptsize (-.11)} \\
 \cmidrule(lr){2-14}
 & \multirow{2}{*}{1} & 19.3 & 23.8 & 25.0 & 37.1 & 59.5 & 60.3 & 60.3 & 76.0 & 77.0 & 38.9 & 52.4 & 52.9 \\
 & & {\scriptsize (-.04)} & {\scriptsize (+.11)} & {\scriptsize (+.07)} & {\scriptsize (+.00)} & {\scriptsize (-.20)} & {\scriptsize (-.17)} & {\scriptsize (-.06)} & {\scriptsize (-.38)} & {\scriptsize (-.40)} & {\scriptsize (-.06)} & {\scriptsize (-.09)} & {\scriptsize (-.10)} \\
 \cmidrule(lr){2-14}
 & \multirow{2}{*}{2} & 24.1 & 24.4 & 25.2 & 52.2 & 60.7 & 61.6 & 69.9 & 76.2 & 78.2 & 51.1 & 55.3 & 55.7 \\
 & & {\scriptsize (+.06)} & {\scriptsize (+.13)} & {\scriptsize (+.07)} & {\scriptsize (-.03)} & {\scriptsize (-.18)} & {\scriptsize (-.12)} & {\scriptsize (-.17)} & {\scriptsize (-.43)} & {\scriptsize (-.38)} & {\scriptsize (-.00)} & {\scriptsize (-.02)} & {\scriptsize (-.02)} \\
 \cmidrule(lr){2-14}
 & \multirow{2}{*}{3} & 27.4 & 27.7 & 27.1 & 65.1 & 67.3 & 67.3 & 79.8 & 81.1 & 81.3 & 58.4 & 58.7 & 58.5 \\
 & & {\scriptsize (+.10)} & {\scriptsize (+.13)} & {\scriptsize (+.09)} & {\scriptsize (-.54)} & {\scriptsize (-.41)} & {\scriptsize (-.45)} & {\scriptsize (-.68)} & {\scriptsize (-.62)} & {\scriptsize (-.49)} & {\scriptsize (-.22)} & {\scriptsize (-.20)} & {\scriptsize (-.20)} \\
\midrule
\multirow{8}{*}{Mistral-7B-v0.3} 
 & \multirow{2}{*}{0} & 32.8 & 33.9 & 29.5 & 69.9 & 78.9 & 79.4 & 84.3 & 86.4 & 86.6 & 60.4 & 62.9 & 63.9 \\
 & & {\scriptsize (+.07)} & {\scriptsize (+.03)} & {\scriptsize (+.05)} & {\scriptsize (-.08)} & {\scriptsize (-.20)} & {\scriptsize (-.17)} & {\scriptsize (-.18)} & {\scriptsize (-.24)} & {\scriptsize (-.22)} & {\scriptsize (-.03)} & {\scriptsize (-.11)} & {\scriptsize (-.11)} \\
 \cmidrule(lr){2-14}
 & \multirow{2}{*}{1} & 26.7 & 30.0 & 27.6 & 52.3 & 70.9 & 72.1 & 76.9 & 82.8 & 82.9 & 47.3 & 59.7 & 60.4 \\
 & & {\scriptsize (+.05)} & {\scriptsize (+.04)} & {\scriptsize (+.07)} & {\scriptsize (-.02)} & {\scriptsize (-.10)} & {\scriptsize (-.07)} & {\scriptsize (+.02)} & {\scriptsize (-.15)} & {\scriptsize (-.21)} & {\scriptsize (-.05)} & {\scriptsize (-.08)} & {\scriptsize (-.07)} \\
 \cmidrule(lr){2-14}
 & \multirow{2}{*}{2} & 30.0 & 30.1 & 28.8 & 60.1 & 72.6 & 75.0 & 79.8 & 84.6 & 85.4 & 56.1 & 60.4 & 60.7 \\
 & & {\scriptsize (+.08)} & {\scriptsize (+.05)} & {\scriptsize (+.06)} & {\scriptsize (-.02)} & {\scriptsize (-.14)} & {\scriptsize (-.10)} & {\scriptsize (-.10)} & {\scriptsize (-.14)} & {\scriptsize (-.19)} & {\scriptsize (-.03)} & {\scriptsize (-.07)} & {\scriptsize (-.07)} \\
 \cmidrule(lr){2-14}
 & \multirow{2}{*}{3} & 31.9 & 32.1 & 30.3 & 69.9 & 76.0 & 77.2 & 84.7 & 86.0 & 86.2 & 60.3 & 62.6 & 63.2 \\
 & & {\scriptsize (+.07)} & {\scriptsize (+.02)} & {\scriptsize (+.08)} & {\scriptsize (-.08)} & {\scriptsize (-.16)} & {\scriptsize (-.13)} & {\scriptsize (-.17)} & {\scriptsize (-.28)} & {\scriptsize (-.23)} & {\scriptsize (-.04)} & {\scriptsize (-.12)} & {\scriptsize (-.13)} \\
\bottomrule
\end{tabular}
\end{center}

\clearpage
\clearpage
\section{Clustering \& Early Decoding Results}

The following tables report the results of the All-but-the-Top experiment described in Section~\ref{sec:re}. Unless stated otherwise, the same experimental setup applies throughout.

\begin{center}
\captionof{table}{Baseline clustering metrics and early decoding Accuracy (\%) across decision-critical layer pairs (no intervention applied). For each layer pair, the top row reports Prediction Purity, Prediction ARI, and Early Decoding Accuracy; the bottom row (in parentheses) reports Ground Truth Purity and Ground Truth ARI.}
\label{tab:baseline_window_stacked}
\scriptsize
\setlength{\tabcolsep}{12.0pt}
\begin{tabular}{ll ccc}
\toprule
\textbf{Model} & \textbf{Layer} & \textbf{PD Purity / (GT Purity)} & \textbf{PD ARI / (GT ARI)} & \textbf{Decoding Acc (\%)} \\
\midrule
\multirow{8}{*}{Qwen2.5-3B} 
 & \multirow{2}{*}{$L-2$} & 0.284 & -0.0019 & \multirow{2}{*}{28.38} \\
 & & {\scriptsize (0.275)} & {\scriptsize (-0.0015)} & \\
 \cmidrule(lr){2-5}
 & \multirow{2}{*}{$L-1$} & 0.283 & -0.0002 & \multirow{2}{*}{26.46} \\
 & & {\scriptsize (0.285)} & {\scriptsize (0.0009)} & \\
 \cmidrule(lr){2-5}
 & \multirow{2}{*}{$L$} & 0.425 & 0.1056 & \multirow{2}{*}{26.57} \\
 & & {\scriptsize (0.393)} & {\scriptsize (0.0770)} & \\
 \cmidrule(lr){2-5}
 & \multirow{2}{*}{$\mathbf{L+1}$} & \textbf{0.821} & \textbf{0.5784} & \multirow{2}{*}{\textbf{34.65}} \\
 & & {\scriptsize (\textbf{0.735})} & {\scriptsize (\textbf{0.4101})} & \\
\midrule
\multirow{8}{*}{Qwen3-8B} 
 & \multirow{2}{*}{$L-2$} & 0.282 & 0.0024 & \multirow{2}{*}{26.26} \\
 & & {\scriptsize (0.277)} & {\scriptsize (0.0003)} & \\
 \cmidrule(lr){2-5}
 & \multirow{2}{*}{$L-1$} & 0.285 & 0.0023 & \multirow{2}{*}{26.26} \\
 & & {\scriptsize (0.279)} & {\scriptsize (0.0007)} & \\
 \cmidrule(lr){2-5}
 & \multirow{2}{*}{$L$} & 0.286 & 0.0046 & \multirow{2}{*}{26.77} \\
 & & {\scriptsize (0.284)} & {\scriptsize (0.0031)} & \\
 \cmidrule(lr){2-5}
 & \multirow{2}{*}{$\mathbf{L+1}$} & \textbf{0.978} & \textbf{0.9402} & \multirow{2}{*}{\textbf{82.02}} \\
 & & {\scriptsize (\textbf{0.857})} & {\scriptsize (\textbf{0.6522})} & \\
\midrule
\multirow{8}{*}{Llama-3.1-8B} 
 & \multirow{2}{*}{$L-2$} & 0.327 & 0.0029 & \multirow{2}{*}{26.26} \\
 & & {\scriptsize (0.280)} & {\scriptsize (0.0008)} & \\
 \cmidrule(lr){2-5}
 & \multirow{2}{*}{$L-1$} & 0.327 & 0.0015 & \multirow{2}{*}{26.77} \\
 & & {\scriptsize (0.282)} & {\scriptsize (0.0004)} & \\
 \cmidrule(lr){2-5}
 & \multirow{2}{*}{$L$} & 0.327 & 0.0004 & \multirow{2}{*}{29.29} \\
 & & {\scriptsize (0.277)} & {\scriptsize (-0.0006)} & \\
 \cmidrule(lr){2-5}
 & \multirow{2}{*}{$\mathbf{L+1}$} & \textbf{0.718} & \textbf{0.3860} & \multirow{2}{*}{\textbf{65.25}} \\
 & & {\scriptsize (\textbf{0.644})} & {\scriptsize (\textbf{0.2592})} & \\
\midrule
\multirow{8}{*}{Llama-3.2-3B} 
 & \multirow{2}{*}{$L-2$} & 0.386 & 0.0003 & \multirow{2}{*}{26.26} \\
 & & {\scriptsize (0.283)} & {\scriptsize (-0.0010)} & \\
 \cmidrule(lr){2-5}
 & \multirow{2}{*}{$L-1$} & 0.386 & 0.0001 & \multirow{2}{*}{27.27} \\
 & & {\scriptsize (0.280)} & {\scriptsize (-0.0010)} & \\
 \cmidrule(lr){2-5}
 & \multirow{2}{*}{$L$} & 0.386 & 0.0090 & \multirow{2}{*}{24.95} \\
 & & {\scriptsize (0.286)} & {\scriptsize (0.0008)} & \\
 \cmidrule(lr){2-5}
 & \multirow{2}{*}{$\mathbf{L+1}$} & \textbf{0.619} & \textbf{0.1955} & \multirow{2}{*}{\textbf{50.00}} \\
 & & {\scriptsize (\textbf{0.431})} & {\scriptsize (\textbf{0.0767})} & \\
\midrule
\multirow{8}{*}{Mistral-7B-v0.3} 
 & \multirow{2}{*}{$L-2$} & 0.439 & 0.1155 & \multirow{2}{*}{28.18} \\
 & & {\scriptsize (0.437)} & {\scriptsize (0.1042)} & \\
 \cmidrule(lr){2-5}
 & \multirow{2}{*}{$L-1$} & 0.435 & 0.1174 & \multirow{2}{*}{29.60} \\
 & & {\scriptsize (0.440)} & {\scriptsize (0.1130)} & \\
 \cmidrule(lr){2-5}
 & \multirow{2}{*}{$L$} & 0.440 & 0.1149 & \multirow{2}{*}{34.75} \\
 & & {\scriptsize (0.428)} & {\scriptsize (0.0995)} & \\
 \cmidrule(lr){2-5}
 & \multirow{2}{*}{$\mathbf{L+1}$} & \textbf{0.630} & \textbf{0.3908} & \multirow{2}{*}{\textbf{65.45}} \\
 & & {\scriptsize (\textbf{0.508})} & {\scriptsize (\textbf{0.1943})} & \\
\bottomrule
\end{tabular}
\end{center}
\clearpage

\begin{center}
\captionof{table}{Intervened clustering metrics and early decoding Accuracy across Local Layer Windows ($L-2$ to $L+1$) following top-10 PC removal at $L+1$. Removing top principal components universally collapses spatial cluster structure ($\text{ARI} \approx 0.00$) and early decoding performance across all architectures at $L+1$.}
\label{tab:intervened_window_stacked}
\scriptsize
\setlength{\tabcolsep}{12.0pt}
\begin{tabular}{ll ccc}
\toprule
\textbf{Model} & \textbf{Layer} & \textbf{PD Purity / (GT Purity)} & \textbf{PD ARI / (GT ARI)} & \textbf{Decoding Acc (\%)} \\
\midrule
\multirow{8}{*}{Qwen2.5-3B} 
 & \multirow{2}{*}{$L-2$} & 0.446 & -0.0010 & \multirow{2}{*}{28.38} \\
 & & {\scriptsize (0.275)} & {\scriptsize (-0.0015)} & \\
 \cmidrule(lr){2-5}
 & \multirow{2}{*}{$L-1$} & 0.446 & 0.0035 & \multirow{2}{*}{26.46} \\
 & & {\scriptsize (0.285)} & {\scriptsize (0.0009)} & \\
 \cmidrule(lr){2-5}
 & \multirow{2}{*}{$L$} & 0.446 & 0.0028 & \multirow{2}{*}{26.57} \\
 & & {\scriptsize (0.393)} & {\scriptsize (0.0770)} & \\
 \cmidrule(lr){2-5}
 & \multirow{2}{*}{$\mathbf{L+1}$} & \textbf{0.446} & \textbf{0.0128} & \multirow{2}{*}{\textbf{26.57}} \\
 & & {\scriptsize (\textbf{0.282})} & {\scriptsize (\textbf{0.0002})} & \\
\midrule
\multirow{8}{*}{Qwen3-8B} 
 & \multirow{2}{*}{$L-2$} & 0.295 & 0.0030 & \multirow{2}{*}{26.26} \\
 & & {\scriptsize (0.277)} & {\scriptsize (0.0003)} & \\
 \cmidrule(lr){2-5}
 & \multirow{2}{*}{$L-1$} & 0.296 & 0.0023 & \multirow{2}{*}{26.26} \\
 & & {\scriptsize (0.279)} & {\scriptsize (0.0007)} & \\
 \cmidrule(lr){2-5}
 & \multirow{2}{*}{$L$} & 0.295 & 0.0050 & \multirow{2}{*}{26.77} \\
 & & {\scriptsize (0.284)} & {\scriptsize (0.0031)} & \\
 \cmidrule(lr){2-5}
 & \multirow{2}{*}{$\mathbf{L+1}$} & \textbf{0.296} & \textbf{0.0002} & \multirow{2}{*}{\textbf{26.26}} \\
 & & {\scriptsize (\textbf{0.283})} & {\scriptsize (\textbf{0.0019})} & \\
\midrule
\multirow{8}{*}{Llama-3.1-8B} 
 & \multirow{2}{*}{$L-2$} & 0.598 & 0.0075 & \multirow{2}{*}{26.26} \\
 & & {\scriptsize (0.280)} & {\scriptsize (0.0008)} & \\
 \cmidrule(lr){2-5}
 & \multirow{2}{*}{$L-1$} & 0.598 & 0.0004 & \multirow{2}{*}{26.77} \\
 & & {\scriptsize (0.282)} & {\scriptsize (0.0004)} & \\
 \cmidrule(lr){2-5}
 & \multirow{2}{*}{$L$} & 0.598 & 0.0029 & \multirow{2}{*}{29.29} \\
 & & {\scriptsize (0.277)} & {\scriptsize (-0.0006)} & \\
 \cmidrule(lr){2-5}
 & \multirow{2}{*}{$\mathbf{L+1}$} & \textbf{0.598} & \textbf{0.0103} & \multirow{2}{*}{\textbf{29.60}} \\
 & & {\scriptsize (\textbf{0.299})} & {\scriptsize (\textbf{0.0039})} & \\
\midrule
\multirow{8}{*}{Llama-3.2-3B} 
 & \multirow{2}{*}{$L-2$} & 0.686 & 0.0010 & \multirow{2}{*}{26.26} \\
 & & {\scriptsize (0.283)} & {\scriptsize (-0.0010)} & \\
 \cmidrule(lr){2-5}
 & \multirow{2}{*}{$L-1$} & 0.686 & -0.0003 & \multirow{2}{*}{27.27} \\
 & & {\scriptsize (0.280)} & {\scriptsize (-0.0010)} & \\
 \cmidrule(lr){2-5}
 & \multirow{2}{*}{$L$} & 0.686 & 0.0130 & \multirow{2}{*}{24.95} \\
 & & {\scriptsize (0.286)} & {\scriptsize (0.0008)} & \\
 \cmidrule(lr){2-5}
 & \multirow{2}{*}{$\mathbf{L+1}$} & \textbf{0.686} & \textbf{0.0031} & \multirow{2}{*}{\textbf{26.87}} \\
 & & {\scriptsize (\textbf{0.276})} & {\scriptsize (\textbf{-0.0014})} & \\
\midrule
\multirow{8}{*}{Mistral-7B-v0.3} 
 & \multirow{2}{*}{$L-2$} & 0.429 & 0.0248 & \multirow{2}{*}{28.18} \\
 & & {\scriptsize (0.437)} & {\scriptsize (0.1042)} & \\
 \cmidrule(lr){2-5}
 & \multirow{2}{*}{$L-1$} & 0.435 & 0.0266 & \multirow{2}{*}{29.60} \\
 & & {\scriptsize (0.440)} & {\scriptsize (0.1130)} & \\
 \cmidrule(lr){2-5}
 & \multirow{2}{*}{$L$} & 0.431 & 0.0230 & \multirow{2}{*}{34.75} \\
 & & {\scriptsize (0.428)} & {\scriptsize (0.0995)} & \\
 \cmidrule(lr){2-5}
 & \multirow{2}{*}{$\mathbf{L+1}$} & \textbf{0.417} & \textbf{0.0086} & \multirow{2}{*}{\textbf{27.17}} \\
 & & {\scriptsize (\textbf{0.287})} & {\scriptsize (\textbf{0.0022})} & \\
\bottomrule
\end{tabular}
\end{center}
\clearpage

\begin{center}
\captionof{table}{\textbf{Intervention effects around the isotropy cliff.} The top 10 principal components are either removed or added at layers $L-2$, $L-1$, $L$, and $L+1$, where $L \rightarrow L+1$ denotes the natural isotropy cliff transition. For each layer, the top row reports the Remove condition, and the bottom row (in parentheses) reports the Add condition. The baseline corresponds to the un-intervened model.}
\label{tab:pc_intervention_results}
\scriptsize
\setlength{\tabcolsep}{12.0pt}
\begin{tabular}{ll ccc}
\toprule
\textbf{Model} & \textbf{Layer / Setting} & \textbf{Final Acc.} & \textbf{$\Delta$ Acc.} & \textbf{Mean IsoScore} \\
\midrule
\multirow{9}{*}{Qwen2.5-3B}
 & Baseline & 0.7889 & 0.0000 & 0.0122 \\
 \cmidrule(lr){2-5}
 & \multirow{2}{*}{$L-2$} & 0.7818 & -0.0071 & 0.0167 \\
 & & {\scriptsize (0.7768)} & {\scriptsize (-0.0121)} & {\scriptsize (0.0120)} \\
 \cmidrule(lr){2-5}
 & \multirow{2}{*}{$L-1$} & 0.7859 & -0.0030 & 0.0143 \\
 & & {\scriptsize (0.7828)} & {\scriptsize (-0.0061)} & {\scriptsize (0.0120)} \\
 \cmidrule(lr){2-5}
 & \multirow{2}{*}{$L$} & 0.7727 & -0.0162 & 0.0163 \\
 & & {\scriptsize (0.7616)} & {\scriptsize (-0.0273)} & {\scriptsize (0.0104)} \\
 \cmidrule(lr){2-5}
 & \multirow{2}{*}{$\mathbf{L+1}$} & \textbf{0.4101} & \textbf{-0.3788} & \textbf{0.0344} \\
 & & {\scriptsize (0.7828)} & {\scriptsize (-0.0061)} & {\scriptsize (0.0098)} \\
\midrule
\multirow{9}{*}{Qwen3-8B}
 & Baseline & 0.8556 & 0.0000 & 0.0193 \\
 \cmidrule(lr){2-5}
 & \multirow{2}{*}{$L-2$} & 0.8505 & -0.0051 & 0.0251 \\
 & & {\scriptsize (0.8444)} & {\scriptsize (-0.0112)} & {\scriptsize (0.0179)} \\
 \cmidrule(lr){2-5}
 & \multirow{2}{*}{$L-1$} & 0.8525 & -0.0031 & 0.0227 \\
 & & {\scriptsize (0.8424)} & {\scriptsize (-0.0132)} & {\scriptsize (0.0186)} \\
 \cmidrule(lr){2-5}
 & \multirow{2}{*}{$L$} & 0.8515 & -0.0041 & 0.0208 \\
 & & {\scriptsize (0.8404)} & {\scriptsize (-0.0152)} & {\scriptsize (0.0189)} \\
 \cmidrule(lr){2-5}
 & \multirow{2}{*}{$\mathbf{L+1}$} & \textbf{0.8242} & \textbf{-0.0314} & \textbf{0.0453} \\
 & & {\scriptsize (0.8525)} & {\scriptsize (-0.0031)} & {\scriptsize (0.0173)} \\
\midrule
\multirow{9}{*}{Llama-3.1-8B}
 & Baseline & 0.7162 & 0.0000 & 0.0283 \\
 \cmidrule(lr){2-5}
 & \multirow{2}{*}{$L-2$} & 0.7020 & -0.0142 & 0.0363 \\
 & & {\scriptsize (0.7162)} & {\scriptsize (0.0000)} & {\scriptsize (0.0231)} \\
 \cmidrule(lr){2-5}
 & \multirow{2}{*}{$L-1$} & 0.6949 & -0.0213 & 0.0346 \\
 & & {\scriptsize (0.7061)} & {\scriptsize (-0.0101)} & {\scriptsize (0.0229)} \\
 \cmidrule(lr){2-5}
 & \multirow{2}{*}{$L$} & 0.6939 & -0.0223 & 0.0330 \\
 & & {\scriptsize (0.6939)} & {\scriptsize (-0.0223)} & {\scriptsize (0.0236)} \\
 \cmidrule(lr){2-5}
 & \multirow{2}{*}{$\mathbf{L+1}$} & \textbf{0.4535} & \textbf{-0.2627} & \textbf{0.0709} \\
 & & {\scriptsize (0.7091)} & {\scriptsize (-0.0071)} & {\scriptsize (0.0222)} \\
\midrule
\multirow{9}{*}{Llama-3.2-3B}
 & Baseline & 0.5848 & 0.0000 & 0.0198 \\
 \cmidrule(lr){2-5}
 & \multirow{2}{*}{$L-2$} & 0.5737 & -0.0111 & 0.0272 \\
 & & {\scriptsize (0.5879)} & {\scriptsize (0.0031)} & {\scriptsize (0.0153)} \\
 \cmidrule(lr){2-5}
 & \multirow{2}{*}{$L-1$} & 0.5707 & -0.0141 & 0.0271 \\
 & & {\scriptsize (0.5808)} & {\scriptsize (-0.0040)} & {\scriptsize (0.0156)} \\
 \cmidrule(lr){2-5}
 & \multirow{2}{*}{$L$} & 0.5717 & -0.0131 & 0.0250 \\
 & & {\scriptsize (0.5838)} & {\scriptsize (-0.0010)} & {\scriptsize (0.0160)} \\
 \cmidrule(lr){2-5}
 & \multirow{2}{*}{$\mathbf{L+1}$} & \textbf{0.4364} & \textbf{-0.1484} & \textbf{0.0488} \\
 & & {\scriptsize (0.5889)} & {\scriptsize (0.0041)} & {\scriptsize (0.0156)} \\
\midrule
\multirow{9}{*}{Mistral-7B-v0.3}
 & Baseline & 0.6939 & 0.0000 & 0.0293 \\
 \cmidrule(lr){2-5}
 & \multirow{2}{*}{$L-2$} & 0.6455 & -0.0484 & 0.0501 \\
 & & {\scriptsize (0.6960)} & {\scriptsize (0.0021)} & {\scriptsize (0.0205)} \\
 \cmidrule(lr){2-5}
 & \multirow{2}{*}{$L-1$} & 0.6444 & -0.0495 & 0.0519 \\
 & & {\scriptsize (0.6838)} & {\scriptsize (-0.0101)} & {\scriptsize (0.0210)} \\
 \cmidrule(lr){2-5}
 & \multirow{2}{*}{$L$} & 0.6232 & -0.0707 & 0.0537 \\
 & & {\scriptsize (0.6798)} & {\scriptsize (-0.0141)} & {\scriptsize (0.0205)} \\
 \cmidrule(lr){2-5}
 & \multirow{2}{*}{$\mathbf{L+1}$} & \textbf{0.3535} & \textbf{-0.3404} & \textbf{0.0726} \\
 & & {\scriptsize (0.6859)} & {\scriptsize (-0.0080)} & {\scriptsize (0.0204)} \\
\bottomrule
\end{tabular}
\end{center}

\section{Qwen3 recovery}

The intervention removes the cluster structure at the cliff: in the intervened run, both ARI and early-decoding accuracy stay at their baseline level for several layers beyond the transition, whereas in the original model they jump directly at $L{+}1$. This damage is not permanent. In Qwen3-8B, clusters re-emerge around layer 28, about five layers later than in the original model, and early-decoding accuracy returns to a similar level as in the original model (about 0.8). The largest impact of the intervention still occurs at the cliff, while the final accuracy is largely preserved.

\begin{figure}[h]
\centering
\includegraphics[width=\columnwidth]{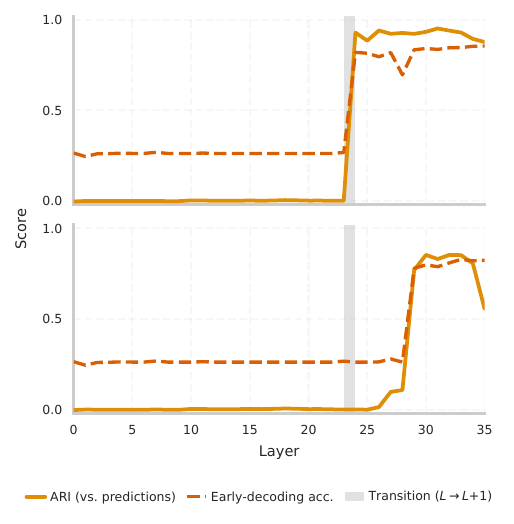}
\caption{Layer-wise ARI (clusters vs.\ model predictions) and early-decoding accuracy for Qwen3-8B. Top: original model; bottom: with the intervention applied at layer $L{+}1=24$. The grey band marks the transition $L \rightarrow L{+}1$ (layers 23$\rightarrow$24).}
\label{fig:qwen3_clean_recovery}
\end{figure}

     \clearpage

\section{2D Projections}
\label{appendix:more-plots}

\makebox[\textwidth][c]{%
    \includegraphics[
        width=\textwidth
    ]{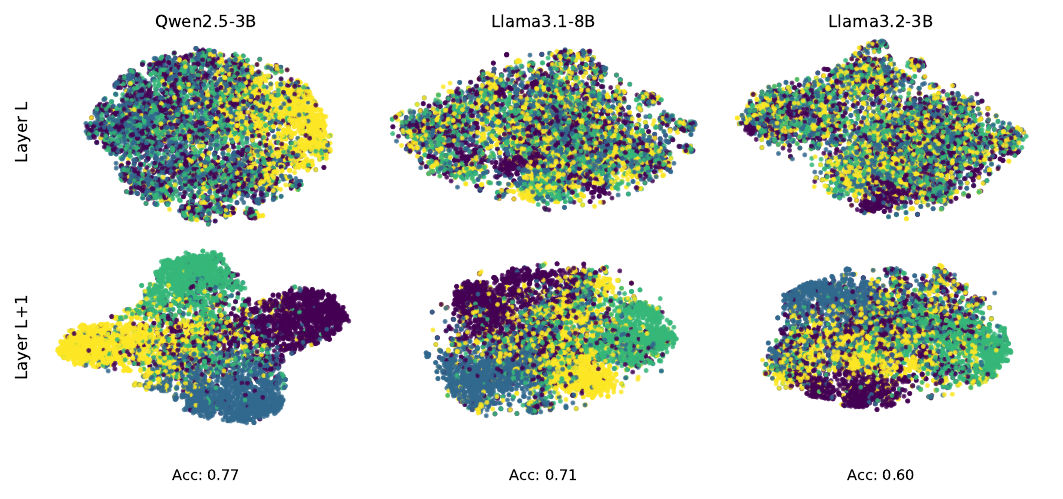}%
}

\noindent

\parbox{\textwidth}{%
    \captionof{figure}{
        \textbf{Clustering analysis of model activations on OpenBookQA using
        prompt format 0 for 5850 samples.} Low-dimensional t-SNE projections of the residual
        stream show representations grouped by ground-truth answer choices.
        This alignment demonstrates behavioral clustering
        \cite{rimsky-etal-2024-steering}, where semantic distinctions between
        choices are reflected in the geometry of the latent space.
    }
    \label{fig:tsne_all}
}

\vspace{2em}

\makebox[\textwidth][c]{%
    \includegraphics[
        width=\textwidth
    ]{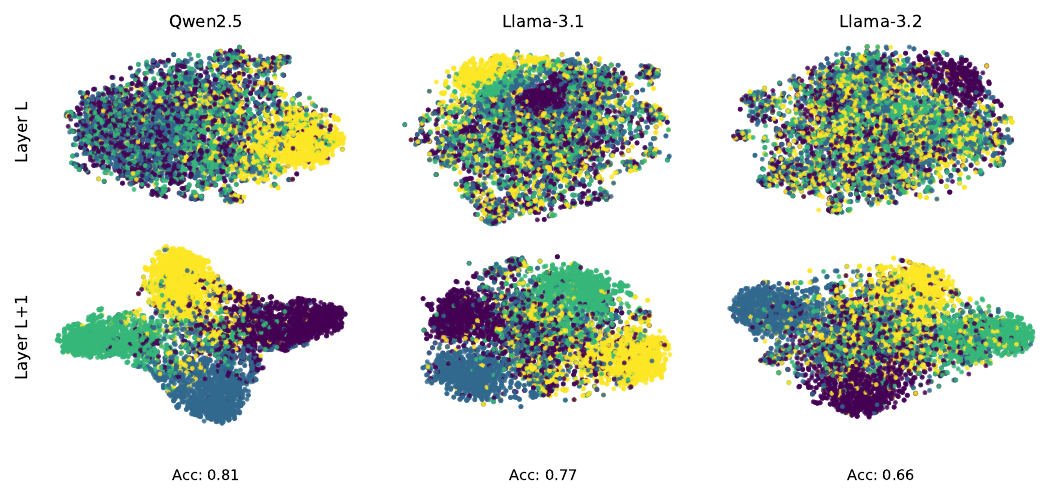}%
}

\parbox{\textwidth}{%
    \captionof{figure}{
        t-SNE projections of hidden representations across adjacent layers
        for OpenBookQA with prompt format 0 and \textbf{4-Shot ICL}.
    }
    \label{fig:tsne_all_4shot}
}

\clearpage


\section{Variance ratio}

\begin{table}[htbp]
\centering
\caption{PCA Cumulative Variance at Cliff Layers}
\label{tab:pca_variance_cliff_layers}
\small
\renewcommand{\arraystretch}{1.15}
\begin{adjustbox}{width=\columnwidth}
\begin{tabular}{l c cc}
\toprule
\textbf{Model} & \textbf{Layer} & \textbf{PC1--10 Variance} & \textbf{PC1--50 Variance} \\
\midrule
\multirow{2}{*}{Qwen3-8B} 
 & Layer 23 & 37.51\% & 63.42\% \\
 & Layer 24 & 62.74\% & 78.90\% \\
\midrule
\multirow{2}{*}{Llama-3.1-8B} 
 & Layer 16 & 33.42\% & 60.34\% \\
 & Layer 17 & 42.97\% & 66.59\% \\
\midrule
\multirow{2}{*}{Llama-3.2-3B} 
 & Layer 14 & 38.73\% & 67.14\% \\
 & Layer 15 & 50.11\% & 73.88\% \\
\midrule
\multirow{2}{*}{Mistral-7B-v0.3} 
 & Layer 18 & 39.68\% & 65.21\% \\
 & Layer 19 & 45.09\% & 68.51\% \\
\midrule
\multirow{2}{*}{Qwen2.5-3B} 
 & Layer 26 & 57.28\% & 76.99\% \\
 & Layer 27 & 72.86\% & 86.06\% \\
\bottomrule
\end{tabular}
\end{adjustbox}
\end{table}

\section{Models \& Datasets}
\label{appendix:appendix_data_models}
\begin{table}[ht]
\centering
\small
\setlength{\tabcolsep}{10pt} 
\caption{Overview of the Large Language Models evaluated in this study.}
\label{tab:appendix_models}
\begin{tabular}{lll}
\toprule
\textbf{Model} & \textbf{License} & \textbf{Reference} \\
\midrule
Qwen2.5-3B      & Qwen Research       & \cite{yang2024qwen2technicalreport} \\
Qwen3-8B        & Apache 2.0          & \cite{yang2025qwen3technicalreport} \\
Llama-3.1-8B    & Llama 3.1 Community & \cite{grattafiori2024llama3herdmodels} \\
Llama-3.2-3B    & Llama 3.2 Community & \cite{grattafiori2024llama3herdmodels} \\
Mistral-7B-v0.3 & Apache 2.0          & \cite{jiang2023mistral7b} \\
\bottomrule
\end{tabular}
\end{table}

\begin{table}[ht]
\centering
\small
\setlength{\tabcolsep}{5pt} 
\caption{Overview of the datasets used for evaluation.}
\label{tab:appendix_datasets}
\begin{tabular}{l c c l l}
\toprule
\textbf{Dataset} & \textbf{Sample Size} & \textbf{Choices} & \textbf{License} & \textbf{Reference} \\
\midrule
ARC-Easy   & 5,100 & 4    & CC BY-SA 4.0 & \cite{clark2018thinksolvedquestionanswering} \\
HEAD-QA    & 5,390 & 4–5  & MIT          & \cite{vilares-gomez-rodriguez-2019-head} \\
OpenBookQA & 5,850 & 5    & Apache 2.0   & \cite{mihaylov2018suitarmorconductelectricity} \\
MMLU-Pro   & 6,000 & 7    & MIT          & \cite{wang2024mmluprorobustchallengingmultitask} \\
\bottomrule
\end{tabular}
\end{table}
    
\end{document}